\documentclass[10pt, a4paper]{article}
\usepackage{lrec2022} 
\usepackage{multibib}
\usepackage{graphicx}
\usepackage{tabularx}
\usepackage{soul}
\usepackage{titlesec}
\titleformat{\section}{\normalfont\large\bfseries\center}{\thesection.}{1em}{}
\titleformat{\subsection}{\normalfont\SmallTitleFont\bfseries\raggedright}{\thesubsection.}{1em}{}
\titleformat{\subsubsection}{\normalfont\normalsize\bfseries\raggedright}{\thesubsubsection.}{1em}{}
\renewcommand\thesection{\arabic{section}}
\renewcommand\thesubsection{\thesection.\arabic{subsection}}
\renewcommand\thesubsubsection{\thesubsection.\arabic{subsubsection}}

\usepackage{epstopdf}
\usepackage[utf8]{inputenc}

\usepackage{hyperref}
\usepackage{xstring}

\usepackage{color}
\usepackage{appendix}

\title{MultiWOZ-DF - A Dataflow implementation of the MultiWOZ dataset}

\name{Joram Meron, Victor Guimarães } 

\address{Telepathy Labs GmbH \\
         36 Militärstrasse, Zurich, Switzerland  \\
         \{joram.meron, victor.guimaraes\}@telepathy.ai\\}

\abstract{
In this paper we introduce MultiWOZ-DF, a dataflow implementation of the well known MultiWOZ dataset. The implementation demonstrates how to both convert and execute MultiWOZ dialogues as dataflow dialogues. Our goal is to encourage the research community to further investigate the dataflow dialogue paradigm. 
\\ 
\newline 
 }

\begin{document}

\maketitleabstract

\section{Introduction}

Semantic Machines (SM) have introduced the use of the dataflow (DF) paradigm to dialogue modelling, using computational graphs to hierarchically represent user requests, data, and the dialogue history \cite{andreas2020task-oriented}.

Although the main focus of that paper was the SMCalFlow dataset (to date, the only dataset with "native" DF annotations), they also reported some results of an experiment using a transformed version of  the commonly used MultiWOZ dataset \cite{budzianowski-etal-2018-multiwoz} 
into a DF format. 

In this paper, we expand the experiments using DF for the MultiWOZ dataset, exploring some additional experimental set-ups. The code and instructions to reproduce the experiments reported here have been released\footnote{https://github.com/telepathylabsai/OpenDF}.

The contributions of this paper are:
\begin{enumerate}
\item A DF implementation capable of executing MultiWOZ dialogues;
\item Several versions of conversion of MultiWOZ into a DF format are presented;
\item Experimental results on state match and translation accuracy.
\end{enumerate}

\section{The MultiWOZ dataset}
The MultiWOZ dataset \cite{budzianowski-etal-2018-multiwoz} consists of 10K annotated dialogues in 7 domains (hotels, restaurants, trains, attractions, hospitals, police, taxi). The collection of the dataset was done using a Wizard-Of-Oz set-up, where both the {\it user} and the system ({\it agent}) are played by human participants. The user is presented with a set of goals they need to achieve (e.g. book a hotel, restaurant, etc.) as well as a set of constraints for the goals (e.g. the hotel should be in a specific area, in a specific price category etc). The {\it user} has no direct access to the data, and thus needs to communicate with the {\it agent}  to achieve the pre-specified goals. The {\it agent}  accesses the data (e.g. hotel information) through a database interface, and their job is to convert the {\it user}'s requests into database queries, informing the {\it user} when the requests succeeds or fails, supplying the information found by the database search, and prompting the {\it user} for additional information in case the request is incomplete, or when the request can not be satisfied.

Two types of annotations are present in MultiWOZ:
\begin{itemize}
\item Dialogue state - the database query that the agent ran at each turn is saved, and is used as a proxy for the dialogue state - i.e. considered as a representation of the agent's belief regarding the user's wishes at that time point. This represents an accumulation of all the turns until that point, and may depend on the agent's subjective interpretation of the user requests;
\item Dialogue acts - a separate annotation pass was done (by a separate group of annotators), marking what information was actually communicated in each turn  (separately for user and agent).
\end{itemize}

The MultiWOZ dataset has been used by many researchers, and is one of the most commonly used dataset for comparison and benchmarking in dialogue system research.

\subsection{Annotation issues}
Both the annotation format, as well as the actual annotations have gone through several iterations of improvements, underlying the inherent difficulties  in trying to achieve complete and objective annotation of natural dialogues. In this work, we use the MultiWOZ-2.2 version \cite{Multiwoz2.2}, which is considered the cleanest version of the dataset.

Despite this, there are still many issues with this annotation, stemming from the assumption that the agent's database query is an exact and objective description of the dialogue state.

In some cases, the values entered into the database query are clearly wrong (the agent may have mistyped,  missed a value, or surmised a value which was not given yet). In other cases, the state may be ambiguous, which is not supported by the annotation scheme (e.g.  after suggesting a value to the user, it may not be immediately clear if the user accepted it or not). The human agent can easily recover from making a wrong choice in such a case, but the resulting annotation is problematic as far as training and evaluation are concerned.

One specific issue with the design of the data collection process is that using the agent's database queries as a proxy to the user's wishes relies on the assumption that the agents can only see the information they specified in their database query, and therefore are forced to explicitly describe what they are looking for. If, for example, the agents could see all the information all of the time (e.g. if they had a screen big enough to display all the information), they would not be forced to create exact queries.

In practice, there are several situations in the interaction, where the agents see more information than they explicitly asked for, and thus are not forced to write full and exact queries. To see why this is a problem consider the following example interaction:

User: "I want a train leaving at 16:15". 

Agent: " there is a train at 16:30". 

User: "How much does it cost?". 

Agent: "the price is ...".  

The agent typed a query with {\it (leave=16:15)} and then got the information for several trains at or after the specified time, and suggested one of them (at 16:30). The user then asked a question about the suggested train (at 16:30), and the agent is able to see the information without having to change the query. It is difficult to say what is the "correct" annotation for this situation (should the time be changed to 16:30 or not), as there is an inherent ambiguity regarding the user's wish at this point, but using the query as a proxy for the state in this situation is problematic.

\section{Dataflow Dialogues}

The dataflow dialogue approach uses computational graphs, which are composed of a rich set of functions (both general and application specific)
to represent the user requests as rich compositional (hierarchical) executable expressions. 

The computational graphs (also referred to as DF expressions) are used for representing the dialogue state/history. In addition, the DF expressions can be executed, which results in manipulating the dialogue state/history, generating an answer (possibly an error message), and optionally producing some side effects through API's to external services (e.g. updating the user's calendar appointments on an external database).

The prominent features of this paradigm are:
\begin{itemize}
\item The dialogue history is represented as a set of graphs, where each computational graph typically represents one user turn;
\item It has a {\it refer} operation to search over the current and previous computational graphs (as well as external resources) which allows easy look-up and re-use of graph nodes which occurred previously in the dialogue;
\item It has a {\it revise} operation which allows modification and reuse of previous computations
\item It has an exception mechanism which allows convenient interaction with the user (e.g. asking for missing information, and resuming the computation once the information is supplied).
\end{itemize}

These features correspond to essential phenomena in natural conversations (referring to previous turns, modifying previous requests, reacting to wrong information, etc.), which allows the system to effectively handle these kinds of user requests.

DF is an object-oriented approach to dialogue design, encouraging modularization and re-use, and may offer practical advantages in terms of scaling and ease of implementing actual dialogue systems.

One interesting aspect of this design is that state (dialogue history), data and computations are all represented using the same format (well structured graphs), which could be useful for new machine learning graph based models (e.g. methods using graph attention), as it allows the model easy access to the state/data.

\section{MultiWOZ and DF}

Despite the fact that MultiWOZ's annotation is essentially flat intent/entity frames, as opposed to the hierarchical (deep) annotations of SMCalFlow,  \cite{andreas2020task-oriented} showed that a DF approach may still be advantageous for MultiWOZ.  

In an experiment, the MultiWOZ annotations were converted into DF expressions, which were then fed into the same pipeline used for training and evaluating the SMCalFlow seq2seq model.  The results showed a small improvement in the {\it joint goal} (average dialogue-state exact match) and {\it dialogue} (average dialogue-level exact match) metrics, compared to the TRADE \cite{wu-etal-2019-transferable} baseline.

Unfortunately,  SM did not release the code for real execution of the DF expressions, which hinders  examination and further experimentation with this approach.

In this work, we release an executable version of the DF implementation of MultiWOZ (based on OpenDF), which should allow researchers to reproduce the MultiWOZ-DF experiments, as well as to extend the work presented here. 

\subsection{Converting MultiWOZ}

When converting MultiWOZ to the DF format, SM converted a dialogue state to a call to a {\it find} function, which gets as parameters the type of service (domain), as well as the values for specified slots.  Within a dialogue, any turn that initiates a new type of booking is re-annotated as a call to {\it find}. Turns that merely modify some of the slots are re-annotated as {\it revise()} calls.  {\it refer()} calls are used to substitute slot values which do not appear verbatim in the user request (provided the refer is correctly resolved).

While the code for converting the MultiWOZ annotations to the DF format were released by SM, we use a different conversion scheme. The differences are described below.

\subsubsection{DF format}
SM's DF expressions use the S-Expression format (a LISP-like format), while our annotations use a Python-like format (which is the standard in OpenDF). 

\subsubsection{MultiWOZ version}
SM used MultiWOZ version 2.1. As mentioned above we use the cleaner version 2.2.

\subsubsection{Conversion source}
SM's conversion used the annotated {\it dialogue state} as the base of conversion. As mentioned above, the dialogue state annotation is derived from the database query, which is manually written by the human agent to describe the user's wishes at that time (the agent generates one query after getting the user's request). SM's conversion used the "state delta" (simply put, the difference between two successive queries) to decide which slots and values should be included in the DF expression for the current user turn. Note that the state delta does not explicitly state which slots/values came from the user, and which from the agent.

Our conversion, on the other hand, is based on the {\it dialogue act} annotations. In this conversion, the DF expressions include only the slots mentioned in the {\it user's} dialogue acts (which has separate annotation for user and agent turns).

In most cases, the two conversions produce essentially similar results (although in different formats). 
However, when information from the agent is incorporated into the dialogue state, the conversions produce different results. As a concrete example, look at the following dialogue:
\

1. User: "I want a restaurant in the center of town"

2. Agent: "How about restaurant X?"

3. User: "good. I want to book a table on Monday"

4. Agent: ". . ."

\
Looking at the conversion for the second user turn (turn \#3), the converted DF expression contains the following slots:
\begin{itemize}
\item Using the state delta based conversion: {\it (day=Monday, name=Restaurant X)}
\item Using the dialogue acts based conversion: {\it (day=Monday)}
\end{itemize}

Practically, this means that for dialogue state based DF expressions, the translation of the user's text request to a DF expression has to also look at the agent's {\it text} in order to extract relevant agent's slots/values from it (in this example - the restaurant name). 

For dialogue act based DF expressions, the translation needs only to extract the user's slots/values. The information from the agent (described by the dialogue acts of the agent) is incorporated into the graphs not through DF expressions, but rather consumed directly by the execution logic. 
 
This makes the DF expression closer to the user request's surface form, and reduces the load on the seq2seq model. 

In a practical dialogue system set-up, this assumption is reasonable, since at run time the human agent is replaced by a programmed agent, which gives us direct access to its internal state and decisions (see section \ref{oracle} for more detail).

\subsubsection{Conversion scope}
SM's conversion is focused on achieving a dialogue state which matches the manual MultiWOZ state annotation.

Our conversion also tries to approximate the behaviour of a more complete dialogue system (the converted expressions should actually execute the dialogues in a way which is acceptable to users).

In a case when the state delta (used by SM) is empty (e.g. when the user asks for some information, like the price of a train), they produce an empty expression. While this behaviour still results in a correct (unchanged) state, it is not a desirable behaviour for a real dialogue system. 

In our conversion, such cases are converted into domain relevant {\it get-info} expressions (which will then produce an answer with the requested information), when the dialogue act indicates information requests by the user. In addition, our implementation takes care of prompting the user for input and keeping the conversation "alive", rather than passively waiting for the user's next input.

\section{Agent oracle}\label{oracle}

Developing a dialogue system, like any other software development, is a process of iterative improvements: the system is tested on real data, problems are identified, solutions are implemented, followed by more tests, etc.   

Due to the context dependent nature of dialogues, a different response from a new version of the system in one turn may cause the rest of the dialogue to follow a different path. This is not a problem if the system interacts with live users, but during the development phase, repeatedly collecting large sets of human-machine interactions for successive iterations of the system is not practical.

Instead, the collection of user interactions is done once, using either a fixed implementation for the system (agent) part, or a WOZ set-up where another human plays the agent.

Therefore, developers face a problem: how to use a fixed dialogue to test a changing system? More specifically, the problem is how to use the fixed part of the human user (since the agent part {\it can} be re-generated).

The problem is more severe when the dialogue system introduces some non-deterministic behaviour.  For example, when several pieces of information are missing, a fully deterministic system may ask for the information pieces one by one in a fixed  order, while a non fully-deterministic system may ask for the information in random order, either one piece at a time or several together (and use different wordings), or even proactively suggest or recommend some values.

The WOZ set-up, where the agent is played by a human, exacerbates the problem further.

\subsection{Agent oracle}

In our implementation we handle this problem by allowing an {\it agent oracle} to influence the system behaviour. Conceptually, the human agent is considered as a non-deterministic system. 

In the implementation of the agent, we separate the non-deterministic part (the decision what actions to take) from the deterministic part (the execution of the decision). Using this separation, we can run the system in different modes during training (where the given agent annotations act as an oracle influencing the decision) and inference (where a rule based decision can be made).

The system can be switched between one of three {\it 'agent oracle'} modes:

\begin{enumerate}
\item {\it Oracle off}: no oracle is present - the agent behaviour and output are generated by executing the programmed logic of the dialogue, this is the typical mode used when running the final system on unseen dialogues;
\item {\it Full Oracle}: an agent's (human) behaviour and response is available  - output the agent's response as is. This is equivalent to the approach taken by \cite{andreas2020task-oriented};
\item {\it Partial Oracle}: an agent's behaviour and response are available - use the agent's behaviour (specifically, the agent's dialogue acts) to guide the programmed logic, and generate the output programmatically.
\end{enumerate}

The {\it partial oracle} mode may also be used to standardize the dialogues (by replacing the human agent's language by the response programmatically generated by the programmed logic), and thus make the seq2seq translation model's work easier. We plan to explore this further in the future.

Using the oracle makes it possible to have "live" interactions with the dialogue system - the dialogues are actually executed by the system, generating DF graphs corresponding to the actual dialogue path. In other words, this means the annotation of MultiWOZ can be extended by complete execution graphs, which could be used for learning and evaluating various models.

Although the oracle's policy (action selected given a specific state) may not always match the policy of the implemented system, the graphs are (hopefully) similar enough to allow the learning of useful inference models.

\section{Conversion alternatives}

We have experimented with two annotation schemes for the DF expressions representing MultiWOZ turns.

\subsection{Full expressions}
This style of DF expressions is closer to the original SM annotations, making direct use of the "raw" {\it revise()} function, which requires the explicit inclusion of "formal" parameters (such as empty type constraints).

\subsection{Simplified expressions}
Similar to the work in \cite{simplifying}, we also implemented a simplified version of the annotation, where the expressions use a modified version of the {\it revise()} function, resulting in shorter, more natural looking expressions, which are closer to the surface natural language request. The simplified expressions were originally introduced to both improve understandability (less relevant in this context) and help the natural language to DF expression translation model.

\subsection{Expression scope}
As mentioned previously, our focus was to make a functional DF dialogue system which can demonstrate how this paradigm works, while using a well known domain. While the goal of SM's experiment was to maximize the match with the annotated state match, our implementation tried to include some additional aspects. For example, when the user asks for information, and the answer does not change the state (e.g. after a restaurant was chosen, the user asks: {\it "Is it expensive?"}) the SM conversion would produces an empty expression. In our conversion, we actually want the system to respond to the answer with a question, so this will be converted to {\it "get\_restaurant\_info(pricerange)"}.

On top of the full and simplified expression alternatives, a flag - {\it omit\_get\_info} - can be used to omit these {\it get\_info} expressions from the conversion.

\section{Metrics}
\begin{figure}
\includegraphics[width=0.5\textwidth]{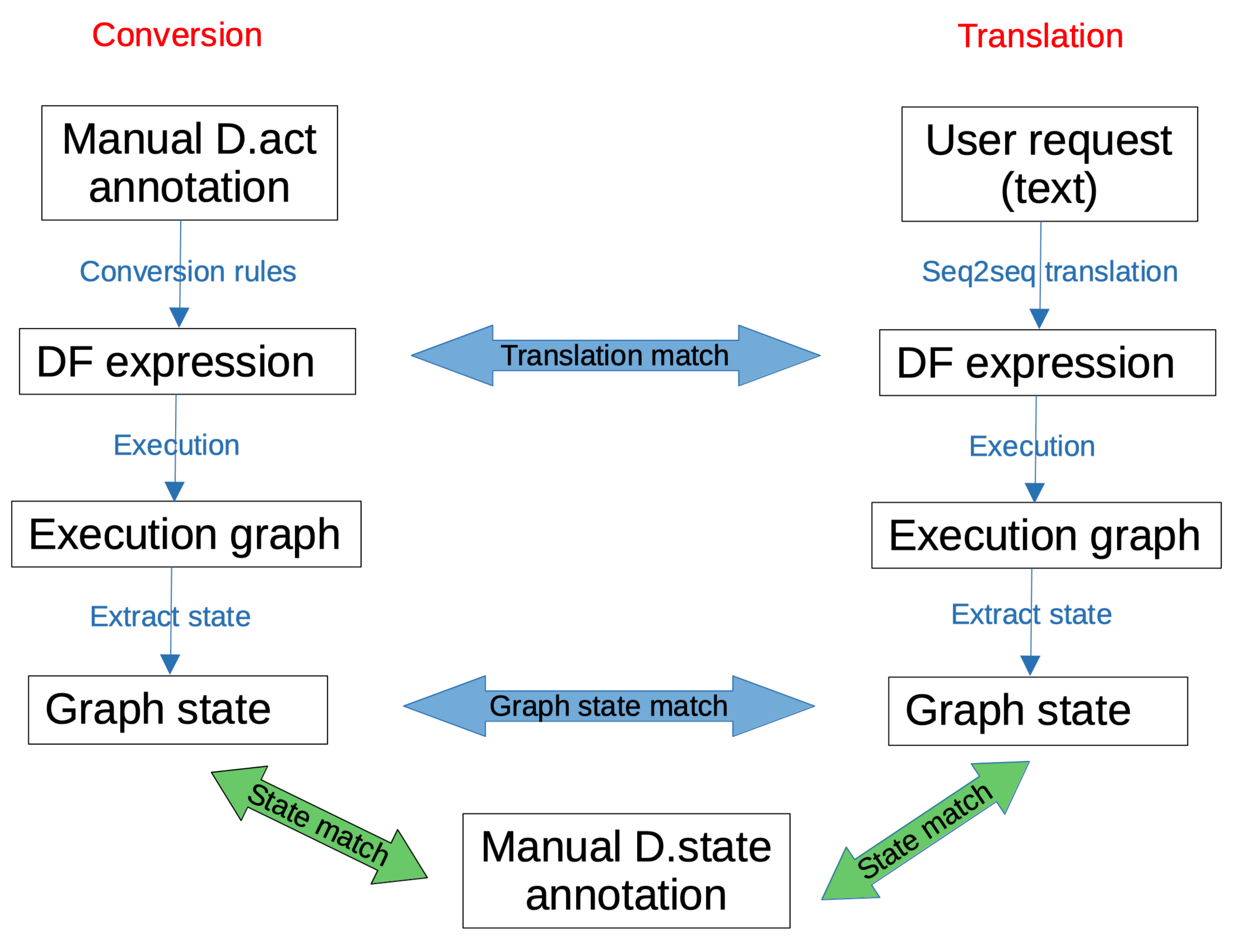}
\caption{\label{compare} Overview of comparisons and metrics used in our experiments. Blue arrows indicate exact match, green arrows indicate lenient match.}
\end{figure}

In this section we describe the metrics used in our experiments. Figure \ref{compare} gives an overview of the way these metrics are used.

\subsection{Translation match}
This metric compares the translation of the users' natural language requests to DF expressions. In SM's paper, the {\it exact match} metric is used. 

\subsection{State match}
This metric compares the dialogue state annotations of MultiWOZ 2.2, at each dialogue turn, to a dialogue state collected from the execution graphs for the same turn. 

The collection of the state from the graph is done in the following way:
\begin{itemize}
\item For complete tasks, we take the complete information - e.g. if a specific hotel was agreed on, all of that hotel's information (which were retrieved from the database and used to instantiate a full hotel sub graph) are added to the collected state. 
\item For incomplete tasks, we take the currently existing fields for the corresponding constraint - e.g. if the user requested a hotel in a price range, but a specific hotel was not selected yet, the collected state will only include the price range field.
\end{itemize}
Our DF conversion is based on the {\it dialogue act} annotations of MultiWOZ 2.2. During dialogue execution, we perform a comparison of MultiWOZ's manual {\it dialogue state} annotation with the dialogue state collected from the execution graph (as described above). This means that we execute the dialogue {\it acts}, but compare the result to the dialogue {\it state}, which is bound to cause mismatch when the two separate manual annotations do not agree.

In our current version, for the  test  set , about 45\% of the dialogues have one or more state mismatch, and almost 15\% of turns have a state mismatch.

Taking a small set of dialogues with state mismatch (randomly selected 25 mismatching dialogues from the development set), we carried out a manual analysis of the cause of the mismatch. The result showed that all of these cases are the result of human annotation errors. 

Note that when comparing the manually annotated state (MS) to the graph state (GS), a more lenient state matching is used than when comparing MS to MS. This is due to practical reasons.

The GS has a set of fields, and for each one (assuming it exists) there can be only one value, while for each MS field there can be a list of values. During comparison we check the following:  1. The set of GS fields is a superset of the set MS fields. 2.  For MS field, at least one of the values is in the corresponding GS field.

The motivation for this were the many cases where the state is ambiguous (and the annotation is arbitrary) - the agent made a suggestion, but it's not clear if the suggestion should be considered as accepted by the user (MultiWOZ, unlike SMCalFlow, does not include explicit annotation for accept/reject suggestions). Further work is needed to address this issue in more detail.

\section{Experiments}

The implemented MultiWOZ-DF was in the following experiments (as shown in figure \ref{compare}):

\subsection{Translation accuracy}

The users' natural language requests were translated to DF expressions using the same seq2seq translation pipeline used by \cite{andreas2020task-oriented}. The translation accuracy results are shown in table \ref{trans-match}.

\begin{table}
\centering
\begin{tabular}{ll}
\hline
   & Turn level exact match \\
\hline
Original & 71.8\% \\  
Simplified & 73.5\%  \\ 
Original  omit & 76.5\% \\ 
Simplified  omit & 78.3\%  \\ 

\hline
\end{tabular}
\caption{\label{trans-match}Turn exact match, on test set. (not averaged over several runs)}
\end{table}

The table confirms that simplified annotations are simpler for the seq2seq translation model to learn, despite the fact that they have identical semantics. It also confirms that omitting the added  {\it get\_info} expressions further increases translation accuracy.

\subsection{State match}

Table \ref{state-match} shows the match result between the manual dialogue state annotation and the execution results of the different conversion alternatives.

\begin{table}
\centering
\begin{tabular}{lll}
\hline
  & Dialogue & Turn \\
\hline
Original, Converted & 55.8\% & 87.2\%\\
Simplified, Converted & 56.1\% & 87.8\% \\
Original, Translated & 40.9\% & 77.9\% \\
Simplified, Translated & 42.7\% & 82.1\%\\
\hline
\end{tabular}
\caption{\label{state-match}
State match, full-dialogue level as well as turn level, on the test set.}
\end{table}

\begin{table}[h]
\centering
\begin{tabular}{lll}
\hline
  & Dialogue & Turn \\
\hline
Original & 47.9\% & 72.2\%\\
Simplified & 54.3\% & 76.0\% \\
\hline
\end{tabular}
\caption{\label{graph-state-match}
Graph state match, comparing converted vs. translated execution graphs, full-dialogue level as well as turn level, on the test set.}
\end{table}

Table \ref{graph-state-match} shows the result of comparing the execution graph state between the converted (ground truth) expressions and the translated expressions, for the non-simplified ("original") and simplified conversion alternatives. Note that the comparison metric used here is exact match (as opposed to the more lenient comparison in table \ref{state-match}).

\section{Further Work}

This work explored the application of the DF paradigm to MultiWOZ. MultiWOZ has an intent-and-entities type of annotation, which is a flat (non-hierarchical) annotation. DF expressions are inherently hierarchical, capable to easily accommodate richer annotations, as exemplified in SMCalFlow.

Still, even for flat annotations datasets, DF may offer benefits. On one hand, it could lead to improvement in learning related aspects of the system (e.g. reducing the need for learning capacity through the use of the {\it refer()} and {\it revise()} functions). On the other hand, DF has a potential to simplify the practical aspects of dialogue system design, with its different approach to dialogue flow design (e.g. as compared to systems based on a state-machine dialogue management).

\section{Conclusion}

In this work we presented MultiWOZ-DF, a dataflow implementation of the widely used MultiWOZ dataset.

We also showed several alternative strategies for converting the MultiWOZ dataset to a dataflow dataset, and the effect this can have on translation and execution accuracy.

The code to reproduce this work has been publicly released at  {\tt https://github.com/ telepathylabsai/OpenDF}.

Previous experiments on a dataflow version of MultiWOZ were reported in \cite{andreas2020task-oriented}, but since that system is not publicly available, follow-up work has been limited.

While direct quantitative comparison of that work's result to our is difficult due to different design decisions taken, our main goal was to allow researchers to get hands-on  experience with a functioning dataflow dialogue system using a familiar dataset, and to encourage further work in this direction.

\section{Bibliographical References}\label{reference}

\bibliographystyle{lrec2022-bib}
\bibliography{mwoz_df}

\appendix

\section{Examples of Dataflow expressions}\label{exExpr}

Table \ref{ex-expr} shows some examples of user requests translated to different alternatives of DF expressions: the SM version, our original version (non-simplified), and our simplified version.

\begin{table*}
\centering
\begin{tabular}{lll}
\hline
 source & text \\
\hline
user & I am looking for a particular restaurant called city stop restaurant.\\
SM  & {\tt  (find (Constraint[Restaurant] :name (?= "city stop restaurant")))}\\
Ours, original  & {\tt cont\_turn(raise\_task(FindRestaurant),}\\
&{\tt     \ \  revise(old=Restaurant??(), newMode=overwrite,}\\
& {\tt    \ \ \ \  new=Restaurant?(name=LIKE(Name(city stop restaurant)))))}\\
Ours, simplified  & {\tt revise\_restaurant(name=city stop restaurant)}\\
\hline
user &  Yes, I would. Please reserve a table for 4 at 17:30 on Thursday.\\
SM  & {\tt (ReviseConstraint :new (Constraint[Restaurant]}\\
& {\tt \ \  :book-day (?= "thursday")}\\
& {\tt \ \ :book-people (?= "4") :book-time (?= "17:30"))}\\
& {\tt \ \ :oldLocation (Constraint[Constraint[Restaurant]])}\\
&{\tt \ \ :rootLocation (roleConstraint \#(Path "output")))}\\
Ours, original  & {\tt revise(old=RestaurantBookInfo?(), newMode=overwrite,}\\
& {\tt \ \  new=RestaurantBookInfo(bookday=thursday,}\\
& {\tt \ \ \ \  bookpeople=4, booktime=17:30))}\\
Ours, simplified  & {\tt revise\_restaurant(bookday=thursday, bookpeople=4, booktime=17:30)}\\
\hline

\end{tabular}
\caption{\label{ex-expr}
User requests,  and their corresponding conversion to DF expressions in three alternatives: SM's version, our original version, our simplified version.}
\end{table*}

\section{Example dialogue}

\begin{figure*}
\includegraphics[height=0.28\textheight,trim= 1.15in 0.5in 0.5in 0.5in]{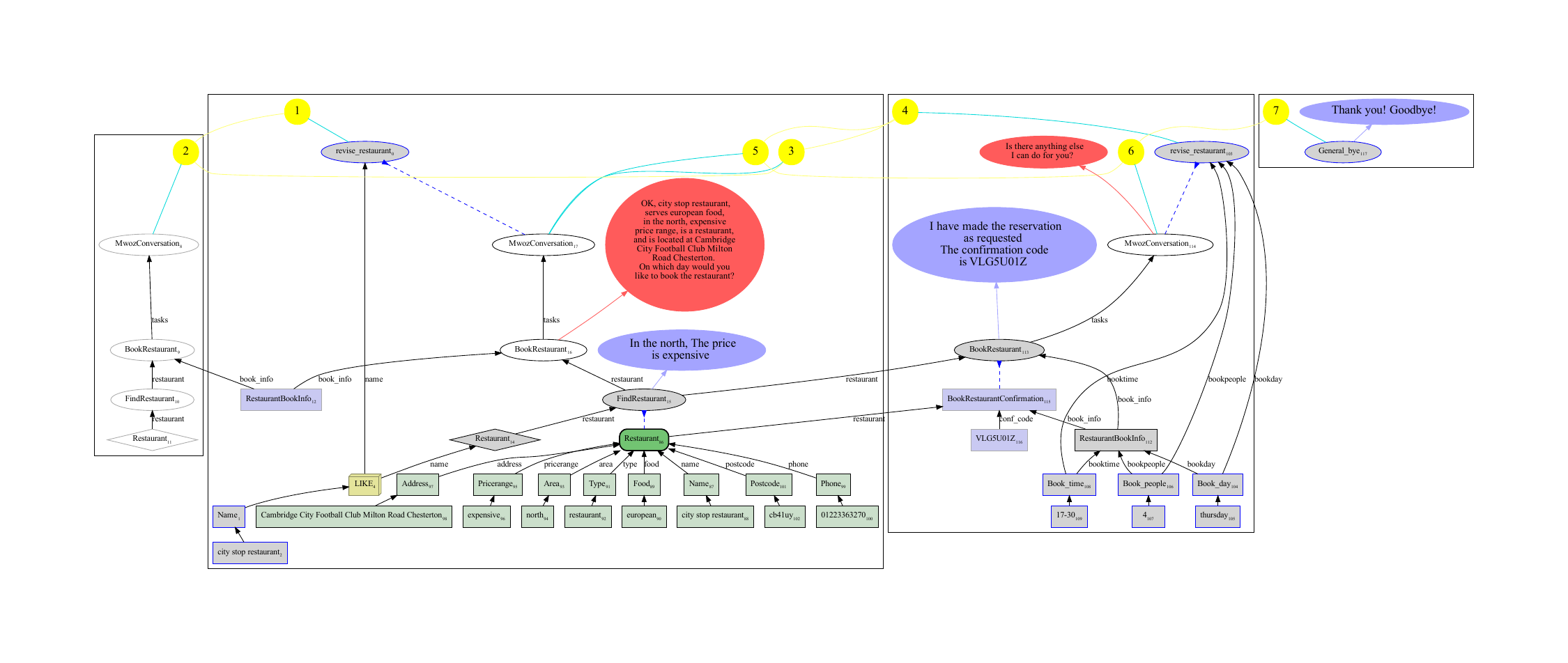}
\caption{\label{ex-dialog} Execution graph for short dialogue (SNG0551), using the simplified expressions.}
\end{figure*}

Figure \ref{ex-dialog}  shows the execution of a short dialogue, and table \ref{ex-txt} shows the user requests, their translation to DF expressions (for the simplified version), and the agent responses for the same dialogue. For the agent response, the table shows both the original agent text ("full oracle"), and the response generated programmatically by the implemented DF nodes.

\begin{table*}
\centering
\begin{tabular}{lll}
\hline
Turn  & source & text \\
\hline
1 & user & I am looking for a particular restaurant called city stop restaurant.\\
1 & DF-expression & {\tt revise\_restaurant(name=city stop restaurant)}\\
1 & full oracle & It is an expensive european restaurant in the north part of town. \\
& &Would you like to make a booking?\\
1 & partial oracle & OK, city stop restaurant, serves european food, in the north, expensive price range, ... \\
& & On which day would you like to book the restaurant?\\ 
\hline
2 & user &  Yes, I would. Please reserve a table for 4 at 17:30 on Thursday. \\
& & I would like the reference number as well, please.\\
2 & DF-expression & {\tt revise\_restaurant(bookday=thursday, bookpeople=4, booktime=17:30)}\\
2 & full oracle & Alright, I've booked a table for 4 at 17:30 on Thursday. Your reference number is VLG5U01Z.\\
2 & partial oracle & I have made the reservation as requested. The confirmation code is VLG5U01Z. \\
& & Is there anything else I can do for you?\\
\hline
3 & user & Thanks, that's all I need. Have a nice day.\\
3 & DF-expression & {\tt General\_bye()}\\
3 & full oracle & You too, enjoy your meal!\\
3 & partial oracle & Thank you! Goodbye!\\
\end{tabular}
\caption{\label{ex-txt}
User requests,  DF expressions, and agent responses (full oracle and partial oracle) for the turns of the short dialogue SNG0551.}
\end{table*}

\end{document}